# Towards Interpretable Deep Extreme Multi-label Learning

Yihuang Kang
*National Sun Yat-sen University*
ykang@mis.nsysu.edu.tw

I-Ling Cheng
*National Chung Hsing University*
chengi428@gmail.com

Wenjui Mao
*National Sun Yat-sen University*
wenjui.mao@gmail.com

Bowen Kuo
*National Sun Yat-sen University*
bowenkuo@outlook.com

Pei-Ju Lee
*National Chung Cheng University*
pjlee@mis.ccu.edu.tw

## Abstract

*Many Machine Learning algorithms, such as deep neural networks, have long been criticized for being "black-boxes"—a kind of models unable to provide how it arrive at a decision without further efforts to interpret. This problem has raised concerns on model applications' trust, safety, nondiscrimination, and other ethical issues. In this paper, we discuss the machine learning interpretability of a real-world application, eXtreme Multi-label Learning (XML), which involves learning models from annotated data with many pre-defined labels. We propose a two-step XML approach that combines deep non-negative autoencoder with other multi-label classifiers to tackle different data applications with a large number of labels. Our experimental result shows that the proposed approach is able to cope with many-label problems as well as to provide interpretable label hierarchies and dependencies that helps us understand how the model recognizes the existences of objects in an image.*

## 1. Introduction

In recent decades, the advance of information technology and ubiquitous computing devices, have fueled the explosive growth of data—the Big Data [1], which is coined by researchers and practitioners to describe this unprecedented phenomenon. Such dramatic increase of data with multimedia contents (e.g. images, audios, videos, and texts) has also facilitated the rapid development of Artificial Intelligence and Intelligent Applications—the applications able to interact with particular environments around them and improve themselves over time. One major reason that triggers the expansion of the intelligent applications is the introduction of Representation Learning [2] in Machine Learning, which allows for automatic data feature extractions with different levels of abstractions. The most popular implementations of the representation learning that cope with multimedia data are Deep Neural Networks (DNNs) [3]. Due to their flexible and deep model architectures, DNNs are theoretically able to create models with most appropriate model capacity/complexity [4] for a given data and thus often outperform other learning algorithms in terms of accuracy of prediction when dealing with massive datasets. DNNs have been very successful in many real-world applications, such as object detection, machine translation, and image captioning [5]–[7]. However, DNNs and many other ensemble machine learning algorithms are often considered "black-box" modeling techniques and thus trained models are relatively hard to interpret. This problem has also raised people's concerns—we cannot assess the models' trust, safety, nondiscrimination, transparency and other ethics. Researchers, practitioners, and policymakers have then started paying attentions on *Machine Learning Interpretability* and *Explainable AI* [8]–[10]—the development of fair, accountable, and transparent intelligent applications.

In this paper, we consider a common application, Multi-label Learning, which involves learning predictive models from annotated/tagged multimedia data with pre-defined labels. Different from typical multi-class classification in supervised machine learning, the target/output variables are a set of binary indicator variables. In Figure 1, we show an example of the multi-label applications that we would like to know what objects are in a picture.

We can see from the picture that there is a dish with annotations/labels automatically identified by online image object detectors. The object detection algorithm actually performed well on recognizing objects in the picture. However, as such algorithms usually learn to identify objects from many pre-defined labels with images, they cannot convey anything but the existences of objects indicated by these labels with corresponding label probabilities. Also, it is very difficult for algorithms to learn from exponential label/target space that involves $2^p$ possible label sets when $p$ is huge—e*Xtreme Multi-label Learning* (XML) [11]–[14] problem coined by researchers in recent years.

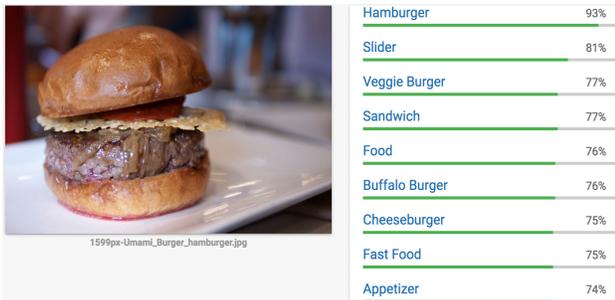

**Figure 1: An image with identified labels**
(Source: Umami_Burger_hamburger.jpg adapted, by Jun Seita, licensed under CC BY 2.0)

Algorithms are proposed to deal with the high-dimensional label space by identifying most relevant subsets of labels, learning label hierarchy, and embedding high-dimensional labels vector onto lower-dimensional ones without sacrificing much information (e.g. label hierarchies and dependencies) within labels [11]–[13]. However, most of these techniques were not originally created for model interpretation purpose and thus we do not know why a model makes a prediction that a particular subset of labels is relevant to an image, text, or tabular data point, which also ignores the information encoded in the labels. Take Figure 1 as an example, we do not know which parts of the image make the algorithm "think" that there is a hamburger. We may accidentally create a plate detector if all hamburger images used to train the model are on plates. Also, we may ignore the knowledge within annotated data, such as dependencies among labels. If the object detector for Figure 1 was trained with labels/ingredients provided by professionals, the model may response with "grilled patty", "ground beef", and even other ingredients like "salt" with high probabilities. It is simply because professional cooks know what ingredients are commonly in the grilled patty, even though these ingredients exhibit significantly visual differences due to diverse ways of cooking—the information/knowledge is virtually encoded within the images and labels.

In this paper, we consider a novel interpretable XML algorithm aiming at predicting most relevant hierarchical label sets as well as providing explanations that help us understand algorithmic decisions given data and labels. Specifically, we propose a two-step XML algorithm that integrates deep non-negative autoencoder with other multi-label classifiers. Take multi-label classification model for image files as the example, the proposed approach may answer questions such as "which part(s) of an image contributes to the label prediction". The non-negative embeddings of the autoencoder learned from large label matrix plays a key role in representing the conceptual label hierarchy and mapping latent label vector into prediction of label sets.

The rest of this paper is organized as follows. We quickly review backgrounds and works related to machine learning interpretability, representation learning, and extreme multi-label learning in Section 2. Our proposed approach, interpretable extreme multi-label learning based on deep non-negative autoencoder, is illustrated in Section 3. In Section 4, we present experimental result when our approach is applied to real-world images with labels. In Section 5, we conclude and summarize our findings.

## 2. Background and Related Work

Machine learning algorithms have been reshaping nearly every corner of our world. From complicated flight planning to everyday grocery shopping, people rely on these algorithms to help make decisions. In recent decades, cheap computation, explosive growth of data, and evolution of deep model architectures [4] have even expanded the capabilities of these algorithms. Deep Learning [3], a powerful marriage of the representation learning and deep neural networks, has proved that distributed (diverse) and deep model architecture can help us identify better models/functions with appropriate model capacity to handle complex tasks. Today, these algorithms have been used to solve many human perceptual tasks, such as visual perception [5] and speech recognition [15]. However, most of these algorithmic models have long been criticized for being "black-boxes", which means it is hard for people to understand these models' decisions.

It is commonly believed that models are supposed to help us make predictions and understand the world of interests [8]. Models unable to be interpreted have been raising people's concerns—"why should I trust the decisions made by the algorithms?". It does not matter a lot when a machine learning algorithm is recommending inappropriate movies to watch, but the stakes are much higher when an algorithm is driving our car or deciding whether a patient in a hospital needs a blood transfusion. People need explanations of why and how algorithms arrive at decisions. Researchers, practitioners, and policy makers have noticed the significance of machine learning interpretability and started working on *Explainable AI* [9], [10], [16] aiming at removing biases and promoting transparency of algorithmic decisions. European Union's new General Data Protection Regulation (GDPR) that took effect in May 2018 have also posed challenges for industry and other organizations to develop algorithms able to provide explanations. People can now ask for reasons of algorithmic decisions that were made about them—*a right to explanation* [17].

In this paper, we define that the machine learning interpretability is "the ability to present model in understandable terms or expressions to a human". We consider the interpretability of a popular real-world application, Extreme Multi-label Learning (XML) [11], [12], [14], [18] for image object detections, which is to learn from tagged images with pre-defined labels. As discussed in Section 1, multi-label learning involves predicting a set of labels in the same domain. By "the same domain", we mean that these label indicators can be transformed into a multi-class target variable with many classes. For example, a movie may belong to many genres (e.g. action, romance, and drama), and $p$ genre indictors can be represented as a single multi-class classification problem with $2^p$ possible classes, which is also how Label Powerset (LP) [19] approach works. Another popular multi-label classification approach is Binary Relevance (BP) and its variations [19], which simply train binary classifier for each label independently and then aggregate/stack predictions of all models. LP and BP are easy to implement, but they often perform poorly, simply because they ignore information within the label matrix (e.g. label dependency and hierarchy) and might suffer severe class imbalance problems when there are too many labels.

The recent growth of annotations on multimedia data aggravates the massive-label problem. Today, an image, text, audio, or video may have thousands or even millions of possible labels/tags/annotations, and therefore it is computationally infeasible for typical multi-label learning algorithms to learn from data with such exponential target label space [11]–[14]. Label compression/reduction techniques, generally classified as tree-based [11], [20] and embedding-based [13], [21]–[23], were proposed to deal with this extreme multi-label classification problem. Tree-based techniques usually learn label hierarchy from training data by recursively partitioning label sets into tree-like structures so as to facilitate the label predictions. On the other hand, embedding-based methods typically assume that latent label dependencies can be encoded in low-rank label matrices, and low-dimensional label target vectors are comparatively easy to predict. However, both kinds of approaches make assumptions that label space can be represented by shallow latent label structures (i.e. tree-like or linear embeddings). Some researchers believe that the assumptions are impractical, because limited and inflexible model structures are usually unable to capture important signals/information in massive datasets [12], [13]. Another important issue about aforementioned XML algorithms is that they were not originally devised to provide model interpretations, which means we are unable to know why these models arrive at a prediction of a label set without further efforts to decipher these models.

Our approach, inspired by Deep Autoencoder [24] and Non-negative Matrix Factorization (NMF) [25], [26], is intended to provide more resilient and interpretable models by taking advantages of learning both label dependency and hierarchy, as well as deep, more expressive [27] model architecture able to create models with higher accuracy and better interpretability. Similar to typical dimensionality reduction techniques like Principal Component Analysis (PCA), the deep autoencoder can be considered a kind of non-linear generalization of PCA but it is able to produce better reconstructions with massive datasets. On the other hand, NMF (and its derivatives) is also a sort of non-probabilistic dimensionality reduction techniques but it is able to provide interpretable and part-based explanations of the original data matrix due to its non-negative nature and the assumption that a non-negative matrix can be represented by low-rank approximations—an object can be represented by additive combinations of some major parts of objects. We here propose a combination of both technique, a deep non-negative autoencoder, which is explainable, flexible, and easy to implement, as it is simply a deep autoencoder with non-negative constraints on embeddings. Another advantage of our approach is that the non-negative latent label vector can also be used with model explainers, such as *Local Interpretable Model-agnostic Explanations* (LIME) [15]. LIME and its variations create explainers by perturbing data instances we would like to explain and learn sparse linear models around them. As LIME is model-agnostic and our approach is able to generate layer-wise non-negative latent label vectors, we can build label hierarchy along with its readable explanations for different multi-label learning applications, such as providing reasons/explanations (i.e. label sets) why the model recognizes that there are some specific objects in a picture.

## 3. Interpretable Extreme Multi-label Learning

We here consider the proposed approach, a two-step interpretable extreme multi-label learning with label compression based on deep non-negative autoencoder. As discussed previously, our proposed non-negative autoencoder is a kind of generalization of the NMF and its non-negative conceptual label sets are relatively easy to interpret. Let $V$ be an $n$ by $p$ non-negative label matrix. A typical NMF is commonly used to find two lower dimensional non-negative matrices $W$ and $H$ such that

$$V \approx WH$$

where $W$ is an $n$ by $k$ basis matrix and $H$ is a $k$ by $p$ coefficient matrix. The goal of the NMF here is to find a low-rank approximation to the label data matrix $V$ by minimizing the Frobenius norm $\|.\|_F$, as the following objective function:

$$\min_{W,H} f(W,H) \equiv \frac{1}{2}\|V-WH\|_F^2, s.t. W \geq 0, H \geq 0$$

where $k < \min(n, p)$ and all elements in $W$ and $H$ are also non-negative. The NMF is a shallow (as opposed to the deep autoencoder) modeling technique and has relatively less expressive power. Instead, we consider implementing a hierarchical version of the NMF based on deep autoencoder with non-negative constraints—a deep non-negative autoencoder, which is here defined to minimize reconstruction error of the original label matrix $V$, as:

$$\|V-V'\|_F^2 = \|V-VH_1H_2...H_LH_L^T...H_2^TH_1^T\|_F^2, s.t. H \geq 0, L > 0$$

where $L$ is the number of encoding-decoding layers and $H_L$ is a non-negative coefficient matrix used to encode/decode the label matrix $V$. The label matrix $V$ can be encoded into intermediate, lower-dimensional, latent label representation matrix $W_L$, as:

$$W_1 = V H_1$$
$$W_2 = W_1 H_2$$
$$\vdots$$
$$W_L = W_{L-1} H_L$$

where $H_L \geq 0$ and $W_L \geq 0$. In Figure 2, we illustrate the autoencoder structure with three fully-connected hidden layers as the example.

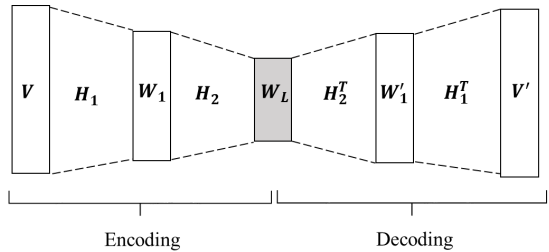

**Figure 2: An example of proposed non-negative autoencoder**

The proposed deep non-negative autoencoder is actually a variation of a hierarchical NMF, but it allows for more flexible and expressive model architectures. In Figure 2, $W_L$ is a latent label representation of the label matrix, and thus can also be used as the lower-dimensional pseudo label output in the following extreme multi-label learning tasks. Besides, due to its non-negative, hierarchical, and part-based nature, the coefficient matrix $H_L$ can be used to explain the label hierarchy.

The next step is to see whether the latent label representations generated by the proposed non-negative autoencoder can help us build extreme multi-label image classifiers as well as provide interpretable hierarchical abstract labels. We consider fine-tuning pre-trained deep

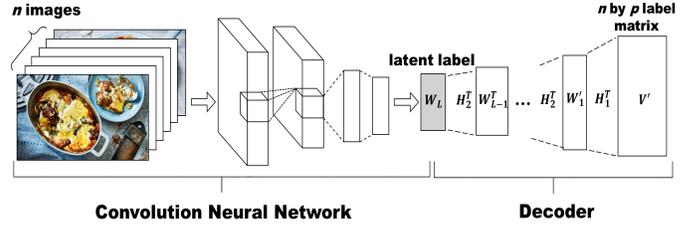

**Figure 3: Interpretable extreme multi-label image classification**

Convolutional Neural Networks (CNN) (e.g. VGG-16 and VGG-19 [5]) with the latent label vectors as the output. Figure 3 shows the network model structure of our proposed approach. Unlike typical multi-label learning tasks, the latent labels here are non-negative real-valued vectors rather than binary indicators, and therefore the output activation function of the CNN is an identity function instead of other non-linear transformation functions (e.g. sigmoid function). The trained decoder plays a crucial role in mapping latent labels back to the reconstruction of the label matrix with original shape for predictions. For model interpretation, on the other hand, the trained autoencoder contains non-negative embeddings (i.e. $H_L$) that map original labels of an image into numeric latent label values. As the embeddings represent the hierarchical label sets that contribute to the final latent label prediction, we can use any model-agnostic explainer algorithms like aforementioned LIME to provide explanations. In the case of image classifier explanations, we are generally interested in top-$N$ superpixels that contribute to the label predictions. In this paper, we consider highlighting superpixels in an image so as to obtain a visual senses of which parts of an image contribute to the prediction of specific label sets.

## 4. Experimental Result

To demonstrate the proposed approach, we collected recipe-ingredient text and dish image data from BBC Food Recipe website [28] (BBC). The recipes without dish images were removed, as we here are only interested in explaining images with label (ingredient) sets at different levels of abstractions. There are total 3,379 recipes with images and 708 unique ingredients. Notice that, for simplicity, here we did not do any further image cropping and rotation. The goal of using the BBC data as the example is to show whether the proposed approach can identify different levels of conceptual label/ingredient sets as the explanations to a given multimedia content with a large number of annotations/labels (image interpretation in this paper). A cooking recipe usually consists of ingredient lists, instructions, and images. And recipes vary with many aspects of human activities, such as cultures and regions [29], [30]. Besides, different recipes may share the same ingredient sets, such as spices, vegetables, stocks, and even cuisines, which form patterns of different levels of label abstractions that may be captured by the aforementioned interpretable multi-label classification models. We empirically used the deep non-negative autoencoder with three fully-connected layers (64 and 16 neurons in two encoding/decoding hidden layers, respectively) to encode/decode the recipe-ingredient label matrix. A pre-trained VGG-16 network was next used as the convolution base to learn the dish images along with the latent labels generated by the autoencoder as the output. Note that we only fine-tuned the top-2 layers of the convolution base and added a new fully connected layer for the pseudo latent label prediction. All the experiments were implemented in R 3.5.1 [31] with Keras [32] and were performed on a computing server with two Intel Xeon CPUs and two NVIDIA Geforce GTX 1080 Ti GPUs.

Figure 4 shows an example of ingredient/label prediction for the dish "Fried Chicken". Here, we provide top-25 most probable ingredients, and the text of matched ingredients are in bold.

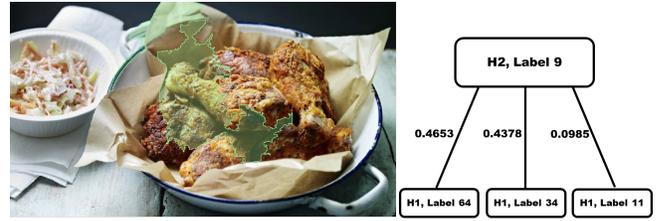

H1, Label 64 : garlic, onion, chicken stock, silverside, garlic bread
H1, Label 34 : garlic, onion, nashi, chicken, guacamole
H1, Label 11 : onion, silverside, chicken stock, charlotte potato, hogget

**Figure 5: Image explanation with label hierarchy for "Fried Chicken"**

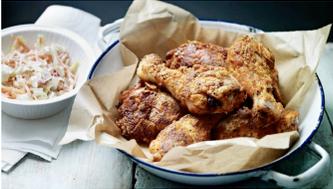

**Figure 4. "Fried Chicken" with actual and predicted ingredients**

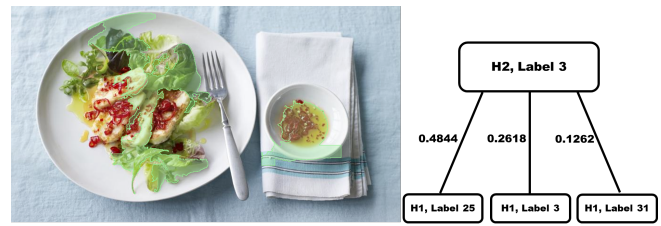

H1, Label 25 : olive oil, chocolate truffle, advocaat, vacherin, fleur de sel
H1, Label 3 : olive oil, dandelion, honeycomb, agar-agar, bone marrow
H1, Label 31 : olive oil, burger, honeycomb, carob, advocaat

**Figure 6: Image explanation with label hierarchy for "Halloumi with quick sweet chilli sauce"**

As there are often "noises" in most of real-word images (e.g. plate in Figure 4), we expect the model would not perform perfectly. We may have created some unknown object detectors as discussed previously. Without clear targets (correctly-labeled and pre-processed images), the model is unable to "perceive" what to predict. On the other hand, without model explanations, we are unable to know why the model made a prediction. Therefore, we are in need of interpretable models or model explainers in such cases. To further justify the prediction, we used the aforementioned LIME to generate the image explanations.

Figure 5 shows an example of the image explanation with its label hierarchy for the dish in Figure 4. The superpixels to explain are highlighted in green. The explainer tells us the highlighted part of the image is most likely the label 9 of the coefficient matrix $H_1$, which consisting of ingredient sets, label 64, 34 and 11 of the coefficient matrix $H_2$. The explanation also suggests that, if we see something resembling to the highlighted part of the image, we could say it is most likely chickens or silversides with ingredient sets made with garlic, garlic bread, onion, guacamole, and so on. Another interesting finding is that, take Figure 5 and Figure 6 as the example, even though the ingredients exhibit large visual differences because of diverse ways of cooking, the explanations with labels can still show us how to visually identify ingredients in a dish, which is usually known by trained professional cooks.

Besides, we also found that some labels never occur in predictions, simply because, in the recipe-ingredient label matrix of the training data, many ingredients/labels occur in only a few recipes/rows, while some ingredients occur in most of the recipes. In the label predictions of BBC data, some dominant ingredients (e.g. salt, eggs, and oils) would always be in the most of the predictions, whereas some essential ingredients (e.g. t-bone steak for recipe "Chargrilled T-Bone Steak") would hardly be found in any predicted label sets—tail label problems [23]. Our approach currently focuses only on building deep model architecture to discover label hierarchy with interpretable abstract labels. It does not explicitly address this issue. It is believed that applying additional sparse decompositions of the label matrix or imposing regularizations/constraints on the weights may solve this problem [12], [23]. We plan to extend the proposed approach so as to deal with this problem in the future.

## 5. Conclusion

We proposed a novel two-step extreme multi-label classification approach that applies deep non-negative autoencoder to the label compression and pseudo label generation of the multi-label learning. The experiment on real-world annotated image data shows that the approach is able to not only build multi-label classification models that cope with a large amount of labels, but also provide layer-wise and part-based explanations to why the model arrived at a label prediction.